# ChatDoctor: A Medical Chat Model Fine-Tuned on a Large Language Model Meta-AI (LLaMA) Using Medical Domain Knowledge

Yunxiang Li [1], Zihan Li [2], Kai Zhang [3], Ruilong Dan [4], Steve Jiang [1], You Zhang [1]

1. Department of Radiation Oncology, University of Texas Southwestern Medical Center, Dallas, USA 2. Department of Computer Science, University of Illinois at Urbana-Champaign, Illinois, USA 3. Department of Computer Science and Engineering, The Ohio State University, Columbus, USA 4. College of Computer Science and Technology, Hangzhou Dianzi University, Hangzhou, CHN

**Corresponding author:** You Zhang, you.zhang@utsouthwestern.edu

## Abstract

### Objective

The primary aim of this research was to address the limitations observed in the medical knowledge of prevalent large language models (LLMs) such as ChatGPT, by creating a specialized language model with enhanced accuracy in medical advice.

### Methods

We achieved this by adapting and refining the large language model meta-AI (LLaMA) using a large dataset of 100,000 patient-doctor dialogues sourced from a widely used online medical consultation platform. These conversations were cleaned and anonymized to respect privacy concerns. In addition to the model refinement, we incorporated a self-directed information retrieval mechanism, allowing the model to access and utilize real-time information from online sources like Wikipedia and data from curated offline medical databases.

### Results

The fine-tuning of the model with real-world patient-doctor interactions significantly improved the model's ability to understand patient needs and provide informed advice. By equipping the model with self-directed information retrieval from reliable online and offline sources, we observed substantial improvements in the accuracy of its responses.

### Conclusion

Our proposed ChatDoctor, represents a significant advancement in medical LLMs, demonstrating a significant improvement in understanding patient inquiries and providing accurate advice. Given the high stakes and low error tolerance in the medical field, such enhancements in providing accurate and reliable information are not only beneficial but essential.

**Categories:** Family/General Practice, Medical Physics, Integrative/Complementary Medicine

## Introduction

The development of instruction-following large language models (LLMs), such as ChatGPT [1], has gained significant attention due to their remarkable success in instruction understanding and human-like response generation. These auto-regressive LLMs [2] are pre-trained on web-scale natural language by predicting the next token and then fine-tuned to follow large-scale human instructions. These models show robust performance on a wide range of natural language processing (NLP) tasks and can generalize to unseen tasks, demonstrating their potential as unified solutions to various problems in natural language understanding, text generation, and conversational artificial intelligence. However, the exploration of such general-domain LLMs in the medical domain remains relatively scarce [3], despite their great potential in revolutionizing medical communication and decision-making [4]. In general, these common-domain models were not trained to capture the medical-domain knowledge specifically or in detail, resulting in models that often provide incorrect medical responses.

By fine-tuning large linguistic dialogue models on data from real-world patient-physician conversations, these models' ability in understanding patients' inquiries and needs can be significantly improved. In addition, to further enhance the models' credibility, a knowledge brain based on online sources such as Wikipedia or offline sources like medical-domain databases can be incorporated into the models to retrieve real-time information to facilitate answering medical questions. The enhanced reliability of such answers is vital for the medical field, as a wrong answer can be detrimental to patients' treatments and well-being. In this study, we investigated the use of these two strategies: model fine-tuning and knowledge brain instillation, to enhance the capability of LLMs to serve as medical chatbots. Since the prevalent ChatGPT model is not open source, we used Meta's public large language model meta-AI (LLaMA) model as the platform for development and evaluation. In detail, we first trained a generic conversation model based on LLaMA, using 52K instruction-following data from Stanford University's Alpaca project [5]. We then fine-tuned the conversation model on our collected dataset of 100K patient-physician conversations from an online medical consultation website (www.healthcaremagic.com). Through extensive experiments, we found that the fine-tuned model by patient-physician dialogues outperforms ChatGPT in terms of precision, recall, and the F1 score [6]. In addition, the autonomous ChatDoctor model, which is able to retrieve the latest online/offline information, can also answer medical questions about relatively new diseases that are not included in the patient-physician training dialogues, for instance, the Monkeypox (Mpox) disease [7,8].

In summary, the ChatDoctor model has the following three main contributions:

1. We established a methodology for fine-tuning LLMs for application in the medical field.

2. We compiled and publicly shared a comprehensive dataset of 100,000 patient-doctor interactions to serve as a training resource for refining the LLM. This dataset includes a wealth of terms, knowledge, and expertise essential for training LLMs in the medical domain. Additionally, we curated and openly shared another dataset consisting of 10,000 patient-doctor conversations from a separate source (www.icliniq.com) to serve as a testing resource for the model. To support and stimulate future advancements in the development of dialogue models in healthcare, we provide public access to all relevant resources such as source codes, datasets, and model weights. These can be found at https://github.com/Kent0n-Li/ChatDoctor.

3. We proposed an autonomous ChatDoctor model that can retrieve online and offline medical domain knowledge to answer medical questions on up-to-date medical terms and diseases, which can potentially reduce the errors and hallucinations of LLMs [9-11].

This article was previously posted to the arXiv preprint server on March 24, 2023.

## Materials And Methods

### Collection and preparation of patient-physician conversation dataset

The initial step in refining our model involved curating a dataset comprising patient-physician interactions. Often, patients describe their symptoms in casual and somewhat superficial language. If we attempted to generate these dialogues synthetically, similar to Alpaca [5], it could lead to over-specific descriptions with limited diversity and relevance to the real world. Hence, we chose to gather authentic patient-doctor conversations, collecting around 100k such interactions from the online medical consultation website, HealthCareMagic. The data were filtered both manually and automatically. Specifically, we automatically filtered out conversations that were too short, most of which did not answer anything of practical significance. And we manually filtered the content of the responses that had errors. To maintain privacy, we erased any information identifying the doctor or the patient and employed LanguageTool to rectify any grammatical errors. This dataset was labeled HealthCareMagic100k, illustrated in Figure *1*. We also sourced roughly 10k additional conversations from another independent online medical consultation site, iCliniq, to test our model's performance. The iCliniq dataset was chosen randomly in a stratified manner to guarantee representation across various medical specialties. It was also made certain that the selected data contained no identifiable patient information, in strict compliance with privacy and ethical standards.

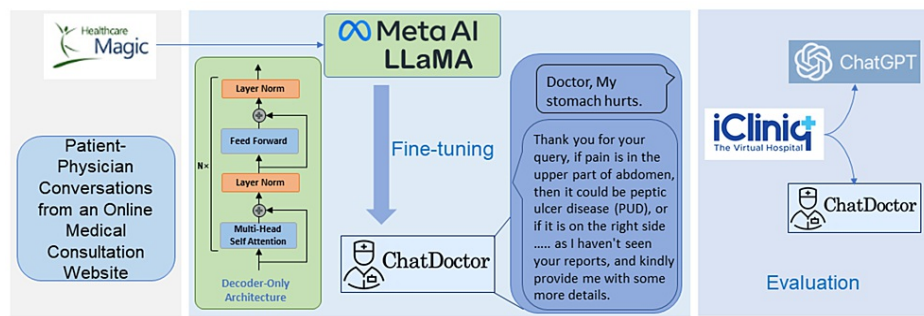

**FIGURE 1: A summary of the process involved in gathering the patient-physician conversation dataset and the steps involved in training the ChatDoctor model.**

### Creation of external knowledge database

LLMs typically predict the next word in a sequence, leading to potential inaccuracies or erroneous responses to questions (hallucinations) [12]. In addition, the model's output can be unpredictable to some extent, which is unacceptable in the medical field. However, the accuracy of these models could be significantly improved if they could generate or assess responses based on a reliable knowledge database, depicted in Figure *2*. Consequently, we curated a database (sample shown in Figure *3*) encompassing diseases, their symptoms, relevant medical tests/treatment procedures, and potential medications. This database serves as an external and offline knowledge brain for ChatDoctor. Continually updatable without requiring model retraining, this database can be tailored to specific diseases or medical specialties. We utilized MedlinePlus to construct this disease database, but other reliable sources can also be used. Additionally, online information sources like Wikipedia can supplement the knowledge base of our autonomous model. It is worth noting that Wikipedia may not be a fully reliable database, but our framework can be easily extended to more reliable online databases such as reputable academic journals.

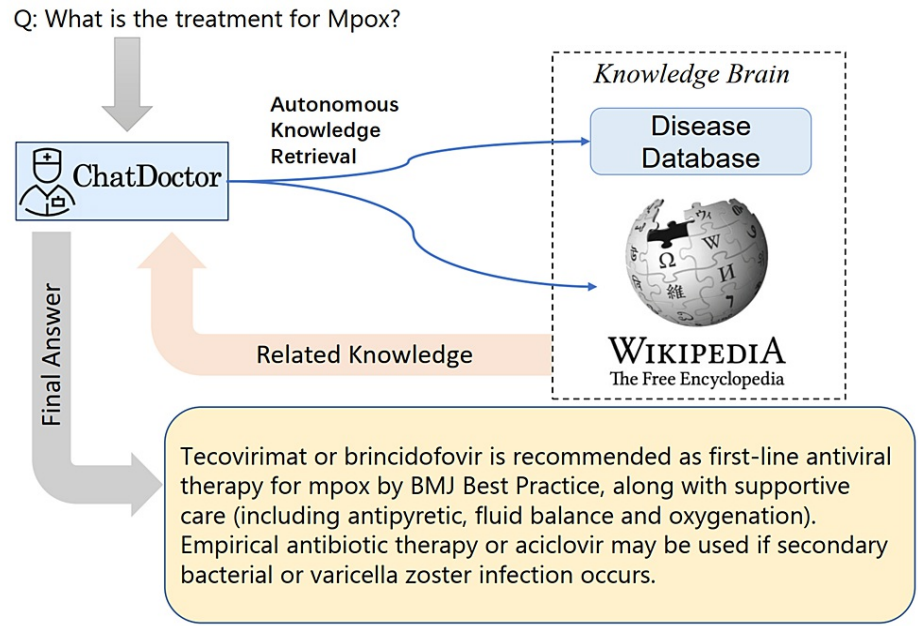

**FIGURE 2: Overview of the autonomous ChatDoctor model based on information retrieval from an external knowledge brain.**

Cureus```
Disease Database

Disease: Appendicitis
Symptoms: Pain in the abdomen, often on the right side. It is usually sudden and gets worse over time. Other symptoms may include: Swelling in the abdomen, Loss of appetite, Nausea and vomiting, Constipation or diarrhea, Inability to pass gas, Low fever
Further test: Abdominal and pelvic CT (Computed Tomography), Abdominal ultrasound, Blood test to check for signs of infection, Urine test to rule out a urinary tract infection
Treatment: Appendectomy, cefotetan (Cefotan), cefotaxime (Claforan), piperacillin and tazobactam (Zosyn), ampicillin and sulbactam (Unasyn), ceftriaxone (Rocephin), cefepime (Maxipime), gentamicin (Garamycin), meropenem (Merrem), ertapenem (Invanz), metronidazole (Flagyl), clindamycin (Cleocin), levofloxacin (Levaquin). In the case of a ruptured appendix, doctors will prescribe an intravenous (IV) antibiotic to treat abdominal infection.
-----------------------------------------------------------------
Disease: Allergic rhinitis
Symptoms: Symptoms that occur shortly after you come into contact with the substance you are allergic to may include: Itchy nose, mouth, eyes, throat, skin, or any area, Problems with smell, Runny nose, Sneezing, Watery eyes. Symptoms that may develop later include: Stuffy nose (nasal congestion), Coughing, Clogged ears and decreased sense of smell, Sore throat, Dark circles under the eyes, Puffiness under the eyes, Fatigue and irritability, Headache.
Further test: Allergy testing, Complete blood count (CBC) testing
Treatment: Antihistamines, Antihistamine nasal sprays, Corticosteroids, Decongestants
-----------------------------------------------------------------
Disease: Malignant otitis externa
Symptoms: Ongoing drainage from the ear that is yellow or green and smells bad. Ear pain deep inside the ear. Pain may get worse when you move your head. Hearing loss, Itching of the ear or ear canal, Fever, Trouble swallowing, Weakness in the muscles of the face.
Further test: Look into the ear for signs of an outer ear infection. The head around and behind the ear may be tender to touch. A nervous system (neurological) exam may show that the cranial nerves are affected. If there is any drainage, the provider may send a sample of it to the lab. The lab will culture the sample to try to find the cause of the infection. To look for signs of a bone infection next to the ear canal, the following tests may be done: CT scan of the head, MRI scan of the head, Radionuclide scan.
Treatment: The goal of treatment is to cure the infection. Treatment often lasts for several months, because it is difficult to treat the bacteria and reach an infection in bone tissue. You will need to take antibiotic medicines for a long period of time. The medicines may be given through a vein (intravenously), or by mouth. Antibiotics should be continued until scans or other tests show the inflammation has gone down. Dead or infected tissue may need to be removed from the ear canal. In some cases, surgery may be needed to remove dead or damaged tissue in the skull.
```

**FIGURE 3: Some samples in our offline disease database consist of symptoms, clinical test/treatment approaches, and medication suggestions.**

### Development of autonomous ChatDoctor with knowledge brain

Armed with the external knowledge brain, i.e., Wikipedia or our custom disease database, ChatDoctor can more accurately answer patient inquiries by retrieving reliable information. Upon establishing the external knowledge brain, we devised a mechanism to enable ChatDoctor to autonomously retrieve necessary information to answer queries. This was accomplished by constructing appropriate prompts to input into the ChatDoctor model. Specifically, we designed keyword mining prompts (Figure *4*) as the initial step for ChatDoctor to extract key terms from patient queries for relevant knowledge search. Based on these keywords, top-ranked information was retrieved from the knowledge brain using a term-matching retrieval system [13]. Given the LLM's word limit (token size), we divided the texts to be read into equal sections and ranked each section by the number of keyword hits. The ChatDoctor model then reads the first N sections (five used in our study) sequentially, selecting and summarizing pertinent information via prompts (Figure *5*). Ultimately, the model processes and compiles all the knowledge entries to generate a final response (Figure *6*). This information retrieval approach ensures patients receive precise, well-informed responses backed by credible sources and can serve as a verification method for responses generated by ChatDoctor





from prior knowledge.

**Prompt for extracting keywords**

A question is provided below. Given the question, extract keywords from the text. Focus on extracting the keywords that can be used to best look up answers to the question.
---
{Question of patient}
---
Provide keywords in the following comma-separated format.
Keywords:

**FIGURE 4: Autonomously extract keywords for information retrieval.**

**Prompt for autonomous knowledge retrieval**

Some information is below.
---
{Relevant content from disease databases or Wikipedia}
---
Select the information that will help to answer the question: {Question of the patient}
Response:

**FIGURE 5: Autonomous information retrieval from the disease database through the prompt.**

**Prompt to make a final answer**

The original question is as follows: {Question of patient}
Based on the information we provided:
---
{Retrieved knowledge}
---
Answer:

**FIGURE 6: Instruct the ChatDoctor to read the retrieved domain knowledge and provide a reliable answer.**

## Model training

We developed the ChatDoctor model using Meta's publicly accessible LLaMA-7B model [14], which uses Transformers with the structure of the decoder only. Despite its relatively modest 7 billion parameters, the LLaMA model exhibits comparable performance to the much larger GPT-3 model (with 175 billion parameters) across several NLP benchmarks. This performance enhancement was achieved by diversifying the training data rather than increasing network parameters. Specifically, LLaMA was trained on 1.0 trillion tokens from publicly accessible data sources like CommonCrawl and arXiv documents. We used conversations from HealthCareMagic-100k to fine-tune the LLaMA model [15] in line with Stanford Alpaca [5] training methodology. The model was first fine-tuned with Alpaca's data to acquire basic conversation skills, followed by further refinement on HealthCareMagic-100k using 6 * A100 GPUs for three hours. The training process followed these hyperparameters: total batch size of 192, learning rate of $2 \times 10^{-5}$, 3 epochs, maximum sequence length of 512 tokens, and a warmup ratio of 0.03, with no weight decay.

## Results

To evaluate the proficiency of the autonomous ChatDoctor model, we tested it using a variety of contemporary medical queries. One of these included a question related to "Monkeypox" (abbreviated as Mpox), as illustrated in Figure 7. Monkeypox was recently designated by the World Health Organization (WHO) on November 28, 2022, making it a relatively novel term. While ChatGPT was incapable of providing a satisfactory response, ChatDoctor, due to its autonomous knowledge retrieval feature, was able to extract pertinent information about Monkeypox from Wikipedia and deliver a precise answer. Similarly, for more





general medical inquiries such as "Otitis," as shown in Figure *8*, ChatDoctor was able to provide a reliable response following the retrieval of relevant knowledge. In another instance, a question about "Daybue," a drug that received FDA approval in March 2023, was accurately addressed by our model after it autonomously retrieved relevant information, demonstrating an advantage over ChatGPT, as shown in Figure *9*.

> **Example of Mpox**
>
> **Q:** How to test for Mpox?
> **ChatGPT:** I'm sorry, but I'm not familiar with the term "Mpox". Could you please provide more information or context about what you're referring to? This will help me provide you with a more accurate answer.
> **ChatDoctor:** Polymerase chain reaction (PCR) testing of samples from skin lesions is the preferred laboratory test. PCR blood tests are usually inconclusive because the virus remains in the blood for only a short time. To interpret test results, information is required on the date of onset of fever, date of onset of rash, date of specimen collection, current stage of rash, and patient age.

**FIGURE 7: Comparison between the ChatGPT and the autonomous ChatDoctor for relatively new medical diseases/terms. The ChatGPT cannot recognize the word Mpox (aka, Monkeypox), while our ChatDoctor can provide the precise answer for the relevant medical tests of Mpox, with the help of the external knowledge brain.**

> **Example of Otitis**
>
> **Q:** How to treat Otitis?
> **ChatGPT:** The treatment of otitis depends on the underlying cause and severity of the condition. Treatment may involve antibiotics, pain relief, ear drops, warm compresses, rest, and prevention measures such as avoiding foreign objects in the ear, keeping ears clean and dry, and protecting ears from loud noises and sudden changes in air pressure. It is important to see a doctor for proper diagnosis and treatment.
> **ChatDoctor:** Treatment depends on the severity of the symptoms. If there is only mild discomfort, then pain relievers such as ibuprofen or acetaminophen can help. Antibiotics may be prescribed if the infection has spread beyond the outer ear. For more serious forms of Otitis, surgery may be needed to remove infected areas of the ear.

**FIGURE 8: Comparison between the ChatGPT and the autonomous ChatDoctor. The ChatGPT provided a more general answer about otitis, while the ChatDoctor provided a more specialized response about the treatments of otitis, with the help of the external knowledge brain.**





> **Example of Daybue**
>
> **Q:** What is Daybue used to treat?
> **ChatGPT:** I'm sorry, but I am not familiar with any medication called "Daybue." It is possible that you have misspelled the name or that the medication is not widely known. Could you please provide me with more information, such as the correct spelling or the condition it is used to treat?
> **ChatDoctor:** Daybue (trofinetide) is used to treat Rett Syndrome.

**FIGURE 9: Comparison between the ChatGPT and the autonomous ChatDoctor. The ChatGPT is unfamiliar with the "Daybue" medication which received approval from the Food and Drug Administration (FDA) in early 2023. The ChatDoctor accurately pointed out the purpose of Daybue (trofinetide), with the help of the external knowledge brain.**

For a quantitative evaluation of ChatDoctor's performance, we utilized questions from the independently sourced iCliniq database as inputs, with the corresponding responses from actual human physicians serving as the benchmark or "ground truth." We compared these with responses generated by both ChatDoctor and ChatGPT. In this evaluation, we employed BERTScore [6] to compute Precision, Recall, and F1 scores for both ChatDoctor and ChatGPT. BERTScore leverages pre-trained BERT to match words in the candidate and reference sentences via cosine similarity, and BERTScore was chosen for its ability to evaluate the semantic similarity between our model's responses and the reference sentences, which we believe is of utmost importance in the medical context. This method of evaluation closely aligns with human judgment at both sentence and system levels. In all three metrics, a higher value denotes a better match. As seen in the results illustrated in Table *1*, the fine-tuned ChatDoctor model outperforms ChatGPT across all three metrics, with specific dialogue examples detailed in Figures *10*-*13*.

|  | ChatGPT | ChatDoctor | P-value |
|---|---|---|---|
| **Precision** | 0.837±0.0188 | 0.8444±0.0185 | $6.66 \times 10^{-195}$ |
| **Recall** | 0.8445±0.0164 | 0.8451±0.0157 | $4.71 \times 10^{-4}$ |
| **F1 Score** | 0.8406±0.0143 | 0.8446±0.0138 | $2.14 \times 10^{-111}$ |

**TABLE 1: Quantitative comparison with BERTScore between ChatDoctor and ChatGPT. The p-values in the table are derived from our paired t-test.**





> **Example 1: ChatGPT vs ChatDoctor.**
>
> **Q:** Hi doctor, I am getting one-sided headache. I do not get them often. They are behind my right eye and temple area. Now this time I am also fighting with heavy cold and cough. It came on late morning today and it hurts more when I cough or shake my head or bend over. Prior to this, I had the same headache about four months ago. At that time it came on midafternoon and lasted till next morning. Prior to that, I had the same headache exactly on the right side and lasted for a day. I am on CCB and Protonix. I am scared of brain tumor. Please help.
>
> **iCliniq:** Hi. Based on what you have told, this is more likely to be due to sinusitis. Sinuses are small air filled cavities in your skull bone. They can get infected and when they are infected, you will get headaches behind eyes. Such headaches will be more common during cold and cough. Brain tumor headache will be on both sides, more in forehead and you will also have vomiting in that case. Based on what you have told, this is more likely due to simple sinus infection instead of brain tumor. CT brain and CT paranasal sinuses.
>
> **ChatGPT:** Hello! I'm sorry to hear you're experiencing these headaches. Based on the location of the pain, it sounds like it may be a tension headache or a migraine. These types of headaches can sometimes be triggered by colds or other illnesses. However, it's always best to get checked out by your doctor to rule out any other underlying conditions. In terms of a brain tumor, that is a very rare cause of headaches. Most people who experience headaches do not have a brain tumor. That being said, if you have any concerning symptoms such as vision changes, weakness, or numbness, it is important to seek medical attention right away. In the meantime, you can try taking over-the-counter pain relievers such as ibuprofen or acetaminophen to help with the headache. It's also important to get plenty of rest and stay hydrated. If the headaches persist or become more frequent, please make an appointment with your doctor. Take care!
>
> **ChatDoctor:** Hi, Thank you for posting your query. First, I would like to reassure you that there is no need to worry about the brain tumor. Also, we should keep migraine apart as well. The most likely cause of your headaches could be sinusitis. This can be confirmed by doing an X-ray of the face. Treatment would consist of using decongestant tablets/ nasal sprays. Best wishes, Chat Doctor.

**FIGURE 10: Example 1: a patient suffering from a unilateral headache expressed concerns about a potential association with a brain tumor. Our ChatDoctor accurately proposed sinusitis as a possible cause for the headache, mirroring the diagnosis provided by the physician from iCliniq. On the other hand, ChatGPT failed to deliver a congruent interpretation regarding the root cause of the one-sided headache.**





> **Example 2: ChatGPT vs ChatDoctor.**
>
> **Q:** Hi doctor, I have one white lump around the tonsil area in my throat. It has been there a good few months now and the area is swollen. Is this cancer?
>
> **iCliniq:** Hi. Cancer cannot be diagnosed by inspection only. To exclude tonsillar stone you have to perform lateral neck X-ray and show me it later on. I advise you to use a mouthwash and gargle your pharynx then expectorate it many times during the day for a week. Then if the white spot still exists you may have to undergo tonsillectomy. Then examine the tonsillar tissue histologically to detect its nature. For more information consult an ENT-otolaryngologist online. Take care.
>
> **ChatGPT:** As an AI language model, I am not qualified to provide medical diagnosis or advice. However, a persistent white lump in the tonsil area that has been present for several months with associated swelling should be evaluated by a medical professional. It could be a benign condition such as a tonsillar cyst or an inflamed lymph node, or something more serious like oral cancer. Therefore, it is recommended that you seek medical attention from a doctor, otolaryngologist (ENT), or head and neck surgeon for further evaluation, diagnosis, and appropriate management.
>
> **ChatDoctor:** Hi dear, thanks for your contact to Chat Doctor. I read your query and understand your concern. As per your complaint it seems that you may be having some kind of enlarged lymph nodes due to some infective cause like reactive hyperplasia or tuberculosis as a possibility. In case if it's not responding to treatment then biopsy should be taken to rule out malignancy. So consult an ENT specialist and get examined. Doctor may order blood test, CT scan or ultrasound of neck to confirm the diagnosis. Doctor may prescribe antibiotics, anti-inflammatory medications depending upon the cause. Hope this will help you for sure. Contact further if follow up needed. Best regards.

**FIGURE 11: Example 2: a patient reported having a white lump in their throat for several months and expressed concerns about potential cancer. All three entities, iCliniq, ChatGPT, and ChatDoctor suggested that the patient could be dealing with abnormally enlarged lymph nodes. Both iCliniq and ChatDoctor additionally recommended that a biopsy and radiological diagnosis would be necessary if initial treatments proved unsuccessful. However, ChatGPT's response was limited to advising the patient to consult with an Ear, Nose, and Throat (ENT) specialist.**





> **Example 3: ChatGPT vs ChatDoctor.**
>
> **Q:** Hi doctor,While working out today, I was doing a shoulder press. I was doing it too fast, and at one point, I got a sharp pain in my back, near where my ribs meet my spine. Since then, I am also having very sharp pain when inhaling and exhaling, as well as when rotating my torso or neck. Should I need to go for urgent care? Currently, I am taking Zyrtec, Amino acid supplements for energy and muscle recovery, hormonal IUD.
> **iCliniq:** Hi. It looks like you have muscle spasm or strain. You can use over-the-counter Motrin 600 mg (Ibuprofen) twice or thrice a day with the meal for two to three days and there are topical analgesics available for the gentle massage. It will get better within two to three days. If it does not get relieved, you can see your doctor. Take rest from the shoulder exercise for a week. Get immediate help if you feel that this pain is significantly disturbing breathing activity. For more information consult an internal medicine physician online
> **ChatGPT:** Yes, it is recommended that you seek urgent medical care for your symptoms. The sharp pain you are experiencing when inhaling and exhaling, as well as when rotating your torso or neck, could be a sign of a rib or back injury. It is important to have an evaluation with a healthcare provider to rule out any serious injuries or conditions that could require further treatment or management. Make sure to inform your provider about your current medications and supplements.
> **ChatDoctor:** 1. The symptoms seem to be due to musculoskeletal origin and not related to any organic pathology. 2. Muscles of back, chest wall and abdomen get strained during vigorous physical activity like weight training. 3. This can cause soreness, stiffness and spasms. 4. This is a self limiting condition and will subside over few days time with adequate rest. 5. You may take anti-inflammatory medications such as ibuprofen for relief from your symptoms. 6. For local application, you can apply warm compresses over affected areas. 7. I hope this information has been both informative and helpful for you. Wish you good health.

**FIGURE 12: Example 3: a patient reported experiencing a sharp back pain during exercise, which intensified during breathing and rotation of the torso or neck. The patient was unsure whether urgent medical attention was necessary. ChatDoctor generated a closer answer to iCliniq than ChatGPT.**





> **Example 4: ChatGPT vs ChatDoctor.**
>
> **Q:** Hello doctor, My vision was crystal clear this morning. I just woke up from a nap, and the vision in my left eye is blurry. It has not cleared up in the past half an hour. Twice over the years, this has happened, and it normalized over a few hours, but I am in a jungle now. So, I would like to err on the side of caution. I have had previous retinal detachments.
> **iCliniq:** Hello. Did you get any surgery done for retinal detachment earlier? It may be a recurrent episode of retinal detachment. If you have diabetes or hypertension, that may be a cause of this detachment. It can also happen in vitreous hemorrhage in which you will see a slightly reddish vision and will not clear up. If this is the case, then it will clear up in few weeks. What is the exact cause can be told only on examination. You should rush to the emergency asap. If detachment surgery is required then visual prognosis depends on the time elapsed between episode and surgery. So, you should not delay. There is not much you can do at the moment by yourself. Just seek help from someone to reach a hospital. For more information consult an eye care ophthalmologist online. Take care.
> **ChatGPT:** Hello, I am an AI language model and I cannot provide medical diagnosis or advice. However, based on your description, a sudden change in vision with a history of retinal detachments raises concern for a possible new detachment or other serious eye condition. It is important that you seek medical attention as soon as possible from an eye specialist. It is better to have it evaluated and treated quickly rather than risking permanent vision loss.
> **ChatDoctor:** 1) You should see your ophthalmologist for a thorough examination and evaluation. 2) There are many causes of sudden loss of vision in one eye, including retina problems such as retinal detachment or bleeding. 3) Your history of prior retinal detachment makes me think that there is something wrong with the retina again and it needs prompt treatment. 4) This can be treated by laser surgery or if needed, a vitrectomy (surgical removal of the vitreous gel). 5) I recommend seeing your ophthalmologist soon for a proper evaluation.

**FIGURE 13: Example 4: a patient experienced blurred vision and was particularly concerned about the health of their left eye. Taking into consideration the patient's past medical history of retinal detachment, all three sources—iCliniq, ChatGPT, and ChatDoctor—advised the individual to seek professional consultation with ophthalmologists for a comprehensive assessment and swift treatment. Due to possible limitations in providing medical diagnoses (and advice), ChatGPT did not speculate on the cause of the diminished vision. On the other hand, both iCliniq and ChatDoctor identified the possibility of retinal detachment or bleeding as potential issues.**

## Discussion

The medical LLM, ChatDoctor, which has been fine-tuned on medical data, has extensive potential uses. These range from preliminary patient assessment and automated case adjudication to proactive healthcare measures. Nevertheless, owing to the complex nature of medical information [16], any concealed inaccuracies in diagnoses and health advice could lead to severe outcomes [17]. LLMs are known to occasionally generate fallacious and harmful assertions (hallucinations) about areas beyond their knowledge expertise, potentially causing medical malpractice [18]. To mitigate this, ChatDoctor has been trained using real-world patient-doctor interactions to better understand patients' questions and deliver more knowledgeable responses. To make the model most capable of answering questions about the latest medical terms (which may not be contained in the training dataset), and to introduce additional external references for verification, we also equipped the ChatDoctor model with the ability to autonomously retrieve information from external knowledge brains to provide answers, further enhancing the credibility of the model [19]. Such external knowledge retrieval can be called by inputting pre-configured prompts into the model. In future developments, the internal prior knowledge of the ChatDoctor model (gained through training) and the external knowledge brain can be further combined by training ChatDoctor to select a more trustworthy answer, or merge and fuse both answers or provide alternative opinions.

### Limitations

It is important to emphasize that the current ChatDoctor model is still in the investigation phase and has been developed for academic research only. The actual clinical use is subject to the risk of wrong answers





being output by the model, and the use of exclusively LLMs in medical diagnosis is still plagued by false positives and false negatives for the time being. Additional security measures, including automated reference checking and human expert evaluation, are needed to cross-validate the answers provided by ChatDoctor to flag potentially inaccurate answers and prevent hallucinations. The exact design, development and deployment of such security measures remains an important topic for further research. A more secure application at this stage is the use of LLMs to assist physicians in their face-to-face consultations. Physicians and ChatDoctor work together to ensure not only that the technology is consistent with clinical practice, but also that patient safety is ensured. The evaluation and potential approval of such tools for healthcare-related purposes also needs further investigation.

## Conclusions

With adequate training and online/offline supervision, ChatDoctor can potentially improve accuracy and efficiency in medical diagnosis and reduce the workload for medical professionals. It may also increase access to high-quality medical consultations, especially for patients in underserved regions with limited medical resources. The further developments and applications of ChatDoctor may eventually help to improve patient outcomes and advance medical research.

## Additional Information

### Disclosures


**Human subjects:** All authors have confirmed that this study did not involve human participants or tissue. **Animal subjects:** All authors have confirmed that this study did not involve animal subjects or tissue. **Conflicts of interest:** In compliance with the ICMJE uniform disclosure form, all authors declare the following: **Payment/services info:** This work was supported by the National Institutes of Health (Grant No. R01 CA240808, R01 CA258987). **Financial relationships:** All authors have declared that they have no financial relationships at present or within the previous three years with any organizations that might have an interest in the submitted work. **Other relationships:** All authors have declared that there are no other relationships or activities that could appear to have influenced the submitted work.